\documentclass[10pt,twocolumn,letterpaper]{article}

 \usepackage{iccv}              

\usepackage[accsupp]{axessibility}
\usepackage{times}
\usepackage{epsfig}
\usepackage{graphicx}
\usepackage{amsmath}
\usepackage{amssymb}

\usepackage{epstopdf}
\usepackage{multirow}
\usepackage{color}
\usepackage[table]{xcolor}
\usepackage{colortbl}
\usepackage{array}
\usepackage{arydshln}
\usepackage{bm}
\usepackage{tabularx}

\usepackage{soul}
\usepackage{url}
\usepackage[utf8]{inputenc}
\usepackage{array}

\usepackage{booktabs}
\usepackage{algorithm}
\usepackage{algorithmic}
\usepackage[switch]{lineno}

\usepackage{colortbl} 
\definecolor{tabletitle}{gray}{.8}
\definecolor{ours}{gray}{.95}
\definecolor{ggray}{RGB}{127,127,127}
\definecolor{reda}{RGB}{202,0,0}
\definecolor{redb}{RGB}{217,148,143}
\definecolor{myyellow}{RGB}{190,144,0}
\definecolor{mygreen}{RGB}{0,136,51}
\definecolor{myblue}{RGB}{0,102,204}
\definecolor{myPurple}{RGB}{117,107,177}
\definecolor{myOrange}{RGB}{230,85,13}
\definecolor{myTeal}{RGB}{27,158,119}
\newcolumntype{B}{!{\vrule width 1pt}}
\usepackage{pifont}

\usepackage[colorlinks,linkcolor=red,anchorcolor=myblue,citecolor=myblue]{hyperref}

\def\paperID{1522} 
\def\confName{ICCV}
\def\confYear{2025}

\title{Controllable-LPMoE: Adapting to Challenging Object Segmentation via Dynamic Local Priors from Mixture-of-Experts}

\author{Yanguang Sun$^1$, Jiawei Lian$^1$, Jian Yang$^{2}$, Lei Luo$^1$ \thanks{Corresponding author.} \\
$^1$PCA Lab, Nanjing University of Science and Technology, Nanjing, China \\
$^2$PCA Lab, VCIP, College of Computer Science, Nankai University, Tianjin, China \\
{\tt\small \{Sunyg, lianjw\}@njust.edu.cn, csjyang@nankai.edu.cn,  luoleipitt@gmail.com}
}

\begin{document}
\maketitle
\begin{abstract}
Large-scale foundation models provide powerful feature representations for downstream object segmentation tasks. However, when adapted to specific tasks through the full-parameter fine-tuning, the enormous parameters being updated often results in significant computational overhead, creating a bottleneck in training efficiency. Although existing methods attempt to fine-tune frozen models by directly embedding trainable prompts, these prompts lack inherent semantic priors, limiting the adaptability of large-scale models. In this paper, we propose a novel dynamic priors-based fine-tuning paradigm with fewer trainable parameters, dubbed Controllable-LPMoE, which adaptively modulates frozen foundation models by dynamically controlling local priors to enhance fine-grained perception for specific segmentation tasks. More specifically, we construct a lightweight dynamic mixed local priors extractor that captures diverse local priors from input images through heterogeneous convolutions while employing a gating network to dynamically output expert priors required for the subsequent fine-tuning. Furthermore, we design a bi-directional interaction adapter that employs cosine-aligned deformable attention and channel-oriented adaptive scale enhancement to interact and restructure between frozen and trainable features, achieving efficient fine-tuning. Extensive experiments validate the superiority of our \href{https://github.com/CSYSI/Controllable-LPMoE} {Controllable-LPMoE} approach, demonstrating excellent segmentation performance compared to 31 state-of-the-art (SOTA) methods and adaptability to multiple binary object segmentation tasks. 
\end{abstract}    
\section{Introduction}
Binary object segmentation, as a fundamental task in computer vision, has been widely studied and encompasses multiple directions, including camouflaged object detection (COD) \cite{SPMAP,GLCONet,SCO,ZoomXNet,FSEL}, salient object detection (SOD) \cite{MDSAM,VST++,MENet,icon}, polyp segmentation (PS) \cite{CFANet,PraNet11,LSSNet}, skin lesion segmentation (SLS) \cite{SLS1,LBUNet}, shadow detection (SD) \cite{sd1,EVP,rmlanet}, glass detection (GD) \cite{RFENet,GlassNet,RUN}, among others. Over the past few years, numerous deep learning-based methods \cite{EBLNet,MDSAM,FSPNet,FSEL,PopNet,DPU-Former} have been proposed, contributing to substantial advances in this field.

\begin{figure*}[]
	\centering\includegraphics[width=0.97\textwidth,height=4.2cm]{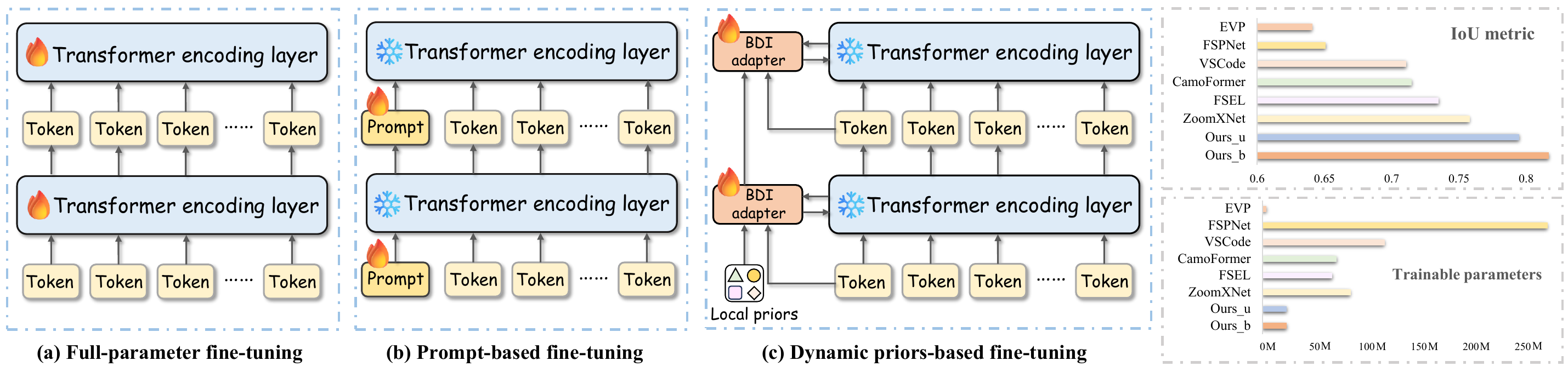}
	\caption{Training paradigms in binary object segmentation tasks: (a) Full-parameter fine-tuning, which updates all model parameters for task adaptation; (b) Prompt-based fine-tuning, which guides learning through trainable prompt embeddings; (c) The proposed dynamic priors-based fine-tuning, which utilizes dynamically controllable local priors to efficiently adapt large-scale models \cite{Beit,uniper}. Additionally, we provide the trainable parameters using different paradigm methods and their IoU accuracy on the challenging COD10K \cite{COD10K} dataset.}
	\label{FIG1}
\end{figure*}
The efficient extraction and encoding of high-quality features from input images is a critical factor for achieving accurate binary object segmentation tasks. Early research \cite{ZoomNet,MENet,FDNet,EMCENet} usually utilizes pre-trained convolutional neural networks ($e.g.$, ResNet50 \cite{ResNet50} with 25.6M parameters or VGG16 \cite{VGG16} with 14.7M parameters) as feature encoders, which have relatively few parameters and can be adapted to specific tasks through full-parameter fine-tuning. Later, Vision Transformers \cite{ViT,Swin,pvt} exhibit strong feature modeling by integrating self-attention and feed-forward networks at each encoding layer. Based on these structures, employing Transformers \cite{FSPNet,VST++,VSCode,ZoomXNet,FSEL} via full-parameter fine-tuning (as depicted in Fig. \ref{FIG1} (a)) has become the mainstream architecture in binary object segmentation tasks, achieving a significant performance breakthrough. However, during the training process, a series of issues appeared consecutively, the most prominent being a sharp increase in memory consumption and a notable decline in training speed, both resulting from the substantial increase in parameters within Vision Transformers \cite{ViT,pvt}. For example, in the challenging COD task, the ZoomXNet \cite{ZoomXNet} approach adopts PVTv2-b5 \cite{pvt}, which has 82M parameters. The trainable parameters for the FSPNet \cite{FSPNet}, FSEL \cite{FSEL}, and CamoFormer \cite{Camoformer} methods are 273.7M, 67.1M, and 71.3M. Similarly, in the VST++ \cite{VST++} model for the SOD task, 112.2M parameters need to be updated, $etc$.

It is evident that when larger-scale Transformer models ($e.g.$, BEiT-L \cite{Beit} with 307M parameters or UniPerceiver-L \cite{uniper} with 302M parameters) with deeper layers and more parameters, which possess stronger modeling capabilities, are used for feature encoding, the feasibility of this fine-tuning paradigm becomes negligible. Recently, the prompt-based fine-tuning paradigm \cite{VPT} has been proposed in visual recognition tasks, which embeds trainable prompts (as shown in Fig. \ref{FIG1}(b)) with few parameters into frozen Transformer layers to enable large-scale models to adapt to specific tasks. Inspired by this, EVP \cite{EVP} and VSCode \cite{VSCode} introduce prompt learning into the binary object segmentation task, acquiring task-specific knowledge through different prompts to fine-tune Transformer \cite{Swin,SegForm} to segment objects in various scenarios. Although the training efficiency of these models \cite{EVP,VSCode} has improved considerably, the segmentation accuracy remains unsatisfactory. As shown in Fig. \ref{FIG1}, on the extremely difficult COD10K \cite{COD10K} dataset, the performance of EVP \cite{EVP} and VSCode \cite{VSCode} is considerably lower than that of ZoomXNet \cite{ZoomXNet} and FSEL \cite{FSEL}, which adopt full-parameter fine-tuning. The reasons for this can be attributed to two aspects: \textbf{1)} it fails to fully leverage the powerful modeling of large-scale models, and \textbf{2)} it is closely related to the inherent properties of prompts. In particular, direct-generated prompts lack prior knowledge, making it challenging to refine object details during iterative training. Furthermore, simply embedding trainable prompts fails to adequately incorporate task-specific knowledge into frozen features, influencing the final segmentation performance.

Taking these reasons into account, we propose a novel dynamic priors-based fine-tuning paradigm in this paper, named Controllable-LPMoE. As illustrated in Fig. \ref{FIG1}, our method involves few trainable parameters ($i.e.$, 23.4M), yet achieves high segmentation accuracy, benefiting from the efficient fine-tuning of large-scale models. Technically, we construct a lightweight dynamic mixed local priors (DMLP) extractor to generate dynamically controllable local priors with task-specific knowledge through multiple heterogeneous convolutions \cite{AC, ASC, WC, DSC} and a mixture-of-experts (MoE) strategy \cite{MoE} from input images for subsequent fine-tuning of large-scale foundation models. Moreover, we design a bi-directional interaction (BDI) adapter to facilitate information transfer between the trainable and frozen features, progressively reconstructing their internal information through iterative updates of cosine-aligned deformable attention and channel-oriented adaptive scale enhancement components. Ultimately, optimized features not only retain powerful universal representations from large-scale models but also incorporate task-specific knowledge, making them efficiently adaptable to segmentation tasks. Extensive experiments on 18 widely-used benchmark datasets from 6 binary object segmentation tasks demonstrate that our Controllable-LPMoE model consistently outperforms 31 state-of-the-art (SOTA) methods. In summary, the main contributions can be summarized as follows:

$\bullet$ A novel dynamic priors-based fine-tuning paradigm is proposed for adapting large-scale models to binary object segmentation tasks through fewer trainable parameters.

$\bullet$ A lightweight dynamic mixed local priors (DMLP) extractor is designed to dynamically capture various local priors using different convolutions and the MoE strategy.

$\bullet$ An efficient bi-directional interaction (BDI) adapter is introduced to reconstruct the representations of trainable and frozen features, leveraging them through interaction.

\begin{figure*}[]
	\centering\includegraphics[width=0.98\textwidth,height=4cm]{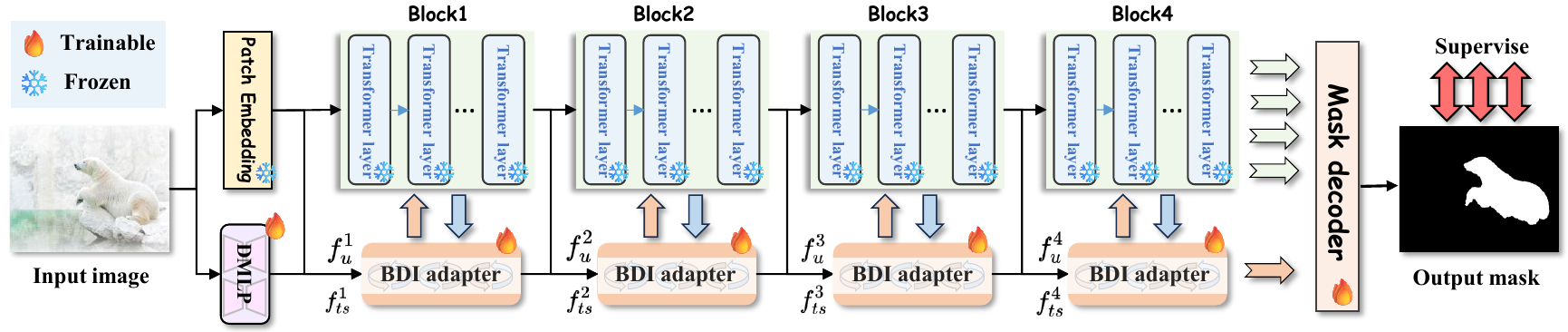}
	\caption{Overview of our Controllable-LPMoE framework. The entire architecture consists of a frozen foundation network \cite{Beit,uniper}, a dynamic mixed local priors (DMLP) extractor, four bi-directional interaction (BDI) adapters, and a mask decoder \cite{mask2}. During training, the proposed Controllable-LPMoE model requires only 23.4M trainable parameters while achieving excellent performance.}
	\label{liucheng}
\end{figure*}

\section{Related Work}

\textbf{Binary object segmentation tasks.} Binary object segmentation is a fundamental research in computer vision, which aims to achieve precise detection and complete segmentation of object regions from input images by constructing a series of frameworks. As a fundamental research, it involves various tasks such as camouflage object detection \cite{FEDER,FSPNet,SegMaR,ZoomXNet,CGCOD}, salient object detection \cite{DSP,VST++,RCNet,UDCNet,SOD11}, medical image segmentation \cite{PraNet11,LSSNet,DCRNet,CFANet}, shadow detection \cite{sdsam,rmlanet,EVP}, glass detection \cite{RFENet,EBLNet,GlassNet}, and so on. Although the categories of objects differ significantly, the design of task-specific architectures is highly similar. The most widely adopted architecture \cite{MEGANet,RFENet,FSEL,icon,semantic} uses a baseline \cite{ResNet50,Swin,pvt} pre-trained on ImageNet \cite{imagenet} to extract initial features, employs the well-designed decoder to generate binary masks, and optimizes the model through a full-parameter fine-tuning paradigm. Excellent performance has been achieved over the past five years \cite{RFENet,dncot,GPONet,M4-SAR,DMINet}. However, with the explosive growth of large-scale models \cite{Beit,dinov2,uniper} and the presence of hundreds of millions of training parameters, this paradigm faces noticeable limitations, such as computational resource constraints, which limit development in the field. Therefore, establishing an efficient fine-tuning paradigm that better adapts large-scale models and maximizes their advantages in feature modeling is meaningful for the future advancement of these tasks.

\textbf{Fine-tuning of large-scale models.} Deep-level structures offer exceptional feature modeling for large-scale foundation models, but also introduce massive amounts of parameters. Recently, some methods \cite{VPT,EVP,VSCode} have explored embedding trainable prompts in large-scale models and integrating them into frozen Transformer layers. By updating a subset of parameters, the model adapts to specific visual tasks. To be specific, VPT \cite{VPT} introduced a small amount of task-specific learnable parameters into the input space for recognition tasks. EVP \cite{EVP} used features from frozen patch embedding and high-frequency components as prompts to fine-tune SegFormer \cite{SegForm} for low-level structure segmentation. OneTracker \cite{prompt3} designed the CMT Prompter and TTP Transformer layer to adapt the Foundation Tracker to downstream RGB+X tracking tasks. VSCode \cite{VSCode} exploited 2D prompts to learn the peculiarities across domain and task dimensions for multimodal SOD and COD tasks. Despite the promising performance of task-specific and multi-task models \cite{prompt3,VPT,EVP,VSCode}, their prompts often lack semantic knowledge and rarely consider the efficient embedding of trainable prompts with frozen structures, which may lead to suboptimal results. 

In this paper, we propose an innovative dynamic priors-based fine-tuning paradigm, called Controllable-LPMoE, which introduces a lightweight dynamic mixed local priors (DMLP) extractor to generate dynamic local priors enriched with task-specific knowledge from input images. Additionally, it constructs a cosine-aligned deformable attention (CDA) for adaptive bi-directional interaction, enabling the efficient fine-tuning of large-scale models to segmentation tasks while utilizing only a few trainable parameters.
\section{Methodology}

\subsection{Overall Architecture}
Fig. \ref{liucheng} illustrates the complete framework of the proposed Controllable-LPMoE method, which consists of four parts: (a) BEiT-L \cite{Beit} / UniPerceiver-L \cite{uniper} foundation encoding model with frozen parameters. (b) Dynamic mixed local priors (DMLP) extractor. (c) Bi-directional interaction (BDI) adapter. (d) Mask decoder \cite{mask2}. For an input image $\mathcal{I}_c$ with size $\mathcal{I}_c\in \mathbb{R}^{3\times H \times W}$, we perform feature encoding in two branches ($i.e.$, task-universal branch, and task-specific branch). The task-universal branch is a large-scale model \cite{Beit,uniper} with frozen parameters that encodes initial features $\{f_u^{i}\}_{i=1}^{5}$, which contain powerful universal representations, each with a size of $\frac{H}{16}\times\frac{W}{16}$. The task-specific branch is a lightweight, trainable DMLP extractor that generates task-specific features $\{f_s^{i}\}_{i=1}^{4}$, each enriched with local priors for the following fine-tuning. Each feature has a spatial resolution of $\frac{H}{2^{i+1}}\times\frac{W}{2^{i+1}}$. Furthermore, we integrate attribute information from both branches using the BDI adapter to generate discriminative features, which are then utilized for binary segmentation through a mask decoder \cite{mask2}.

\subsection{Dynamic Mixed Local Priors Extractor}
\begin{figure}[]
	\centering\includegraphics[width=0.48\textwidth,height=4.7cm]{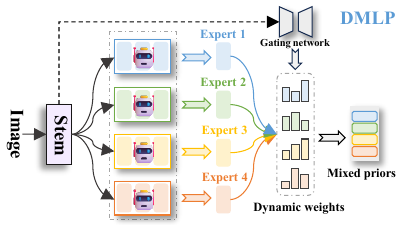}
	\caption{Flowchart of the first stage of our DMLP extractor.}
	\label{DMLP}
\end{figure}

The purpose of our DMLP extractor in the task-specific branch is to capture rich local priors and dynamically control their output for subsequent fine-tuning. During the fine-tuning process, these local priors provide task-specific knowledge for segmentation tasks, while the plentiful spatial details they contain help refine the boundary information of objects. Unlike these spatial priors \cite{Vit-ad,Vit-CO}, the local priors obtained by our DMLP extractor are dynamic and diverse. Technically, given an input image $\mathcal{I}_c$ in the first stage (as shown in Fig. \ref{DMLP}), we employ multiple sets of lightweight heterogeneous convolutions ($i.e.$, depthwise separable convolution \cite{DSC}, atrous convolution \cite{AC}, asymmetric convolution \cite{ASC}, and wavelet convolution \cite{WC}) with different receptive fields to construct four types of local priors $\{\mathrm {E}_{p}^{n}\}_{n=1}^{4}$ that contain task-specific knowledge, which can be written as follows:
\begin{equation}
\begin{split}
        &\mathrm {E}_{p}^{n} = \mathcal{C}_1 ([l_1^{n},l_3^{n},l_5^{n},l_7^{n}]), l_1^{n}=\mathcal{C}_1(\mathsf{stem}(\mathcal{I}_c)), \\
        & l_{2k+1}^{n}=\mathcal{ZC}_{2k+1}^{n}(l_{1}^{n}+l_{2k-1}^{n}), k=1,2,3,
\end{split}
\end{equation}
where $\mathcal{C}_1(\cdot)$, $[\cdot]$, and $\mathsf{stem}(\cdot)$ denote 1$\times$1 convolution, concatenation, and down-sampling operation. $l_{2k+1}^{n}$ and $l_{2k-1}^{n}$ are local priors of the $n$-th type, with different receptive fields. $\mathcal{ZC}_{2k+1}^{n}(\cdot)$ represents $n$-th type of lightweight convolution, with a kernel of size $(2k+1)\times(2k+1)$. Considering the diversity \cite{icon} of local expert priors $\{\mathrm {E}_{p}^{n}\}_{n=1}^{4}$, we propose a dynamic control strategy (DCS) that corrects the proportion of all local priors through dynamic weighting. Specifically, inspired by Mixture of Experts (MoE) \cite{MoE}, we first treat each local prior $\mathrm {E}_{p}^{n}$ as an expert with different knowledge, and then generate a set of dynamic weights \{$w_l^n\}_{n=1}^{4}$ through a gating network based on the input feature $\hat{\mathcal{I}}_c$ ($\hat{\mathcal{I}}_c=\mathsf{stem}(\mathcal{I}_c)$), as shown in: 
\begin{equation}
\begin{split}
        &w_l^n(\hat{\mathcal{I}}_e)= \mathsf{Softmax}(\mathsf{W}_g \hat{\mathcal{I}}_e+\mathsf{b}_g), n=1,2,3,4, \\
\end{split}
\end{equation}
where $\mathsf{W}_g$ and $\mathsf{b}_g$ represent the learnable weight matrix and bias vector from the linear layer.  ``$+$'' denotes the element-wise addition operation.  Furthermore, four experts $\{\mathrm {E}_{p}^{n}\}_{n=1}^{4}$ with different prior knowledge are integrated using dynamic weights $\{w_l^n\}_{n=1}^{4}$ to generate the task-specific feature $f_s^1$, which incorporates enriched local prior semantics. The process is formulated as follows:
\begin{equation}
\begin{split}
        &f_s^1=\mathcal{C}_1(\hat{\mathcal{I}}_c+\sum_{n=1}^{N} w_l^n\otimes\mathrm {E}_{p}^{n}), N=4, \\
\end{split}
\end{equation}
where ``$\otimes$'' is the element-wise multiplication. Our dynamic control strategy (DCS) makes all local priors dynamically controllable, and it can continuously adapt and adjust during the fine-tuning process. The proposed DMLP extractor includes four stages, with each stage using the task-specific feature $\{f_s^{i-1}\}_{i=2}^{4}$ from the previous stage as input and generating the feature $\{f_s^{i}\}_{i=2}^{4}$ through similar operations ($i.e.$, local prior extraction and dynamic integration).

\subsection{Bi-directional Interaction Adapter}
For large-scale models, the simplest and most straightforward manner for applying them to downstream segmentation tasks is to train and fit them directly through full-parameter fine-tuning. However, existing large-scale foundation models \cite{Beit,dinov2,uniper} consist of many layers, and their parameters grow exponentially. Updating all parameters through gradient descent and adapting them to segmentation tasks is extremely time-consuming and labor-intensive. Although some existing methods \cite{EVP, VSCode} attempt to fine-tune large-scale models with frozen parameters by embedding trainable prompts in binary object segmentation tasks, the lack of semantic priors in the generated prompts hinders their ability to effectively leverage the advantages of large-scale pre-trained models for feature modeling.

Considering the above challenges, we design the BDI adapter that includes cosine-aligned deformable attention (CDA) and channel-oriented adaptive scale enhancement (CASE) to exchange information between frozen and trainable features and iteratively update them. On the one hand, the fine-tuning of large-scale models is guided by dynamic local priors with semantic knowledge; on the other hand, rather than simply embedding prompts, the bi-directional interaction between the two features in the BDI adapter facilitates efficient information transfer, further enhancing the fine-tuning performance of large-scale models.

\textbf{Input features.} The BDI adapter leverages the outputs of the frozen model as task-universal features, while incorporating the abundant local semantic priors from the trainable DMLP extractor as task-specific features. Specifically, we evenly divide the frozen encoding model \cite{Beit,uniper} into four blocks, each containing six encoding layers, and universal features $\{f_u^{i}\}_{i=1}^{5}$ ($f_u^{i}\in \mathbb{R}^{\frac{HW}{16^{2}}\times D}$) are obtained from the output of the patch embedding and four blocks. Meanwhile, we flatten and concatenate local features $\{f_s^{i}\}_{i=2}^{4}$ to generate an initial task-specific feature $f_{ts}^1$, that is,
\begin{equation}
\begin{split}
        &f_{ts}^1=\mathsf{flat}([f_s^{2},f_s^{3},f_s^{4}]) \in \mathbb{R}^{(\frac{HW}{8^{2}}+\frac{HW}{16^{2}}+\frac{HW}{32^{2}})\times D}.
\end{split}
\end{equation}
\textbf{Cosine-aligned deformable attention.} For the task-universal feature $f_u^{1}$ and the task-specific feature $f_{ts}^1$, we perform the first knowledge exchange through our CDA mechanism, with the goal of enriching the universal feature with task-specific clues. Technically, we take the normalized feature $\tilde{f}_u^1(\tilde{f}_u^{1}= \mathsf{Norm}({f}_u^1))$ as the primary $query$, where $\mathsf{Norm}(\cdot)$ denotes a layerNorm, and the task-specific feature $\tilde{f}_{ts}^1$ as the auxiliary $value$. To enhance the semantic alignment between $query$ and $value$, we incorporate cosine similarity into the attention weights, enabling the model to focus more on regions with higher relevance, that is,
\begin{equation}
\begin{split}
        &\mathcal{A}_{\phi}^{1}=\mathsf{Softmax}(\Phi(\tilde{f}_u^1)) \otimes \mathsf{Cosine}(\tilde{f}_u^1,\tilde{f}_{ts}^1), \\
        &\mathsf{Cosine}(\tilde{f}_u^1,\tilde{f}_{ts}^1)=\mathsf{Softmax}(\Phi(\frac{\tilde{f}_u^1 \odot \tilde{f}_{ts}^1}{\|\tilde{f}_u^1\| \|\tilde{f}_{ts}^1\|}\otimes \tilde{f}_{ts}^1)), 
\end{split}
\end{equation}
where $\Phi(\cdot)$ denotes a linear layer, $\odot$ is the matrix multiplication operation, $\|\cdot\|$ represents the Euclidean norm. By weighting the input feature, our approach dynamically adjusts the attention weight distribution. Subsequently, we update the internal information of the initial input feature, and the formula can be expressed as:
\begin{equation}
\begin{split}
&\hat{f}_u^{1}=f_u^1+\Phi(\mathcal{A}_{\phi}^{1}\otimes\mathsf{W}_v\tilde{f}_{ts}^1(p_q^{1}+\nabla p_{so}^{1}))\otimes\Psi_ o, \\
\end{split}
\end{equation}
where $p_q^1$ and $\nabla p_{so}^{1}$ represent the 2-$d$ reference point related to the $query$ and the sampling offset \cite{Deform}, while $\Psi_o$ denotes a learnable vector variable initialized to 0, balancing the attention layer's output and the input $query$. 

\begin{table*}[]
\centering
\setlength{\tabcolsep}{5pt}
\renewcommand{\arraystretch}{0.9}
\resizebox*{\textwidth}{57mm}{
\begin{tabular}{c|c|cccc|cccc|cccc|cccc}
\hline \hline
\multirow{2}{*}{Methods} & \multirow{2}{*}{Pub.} & \multicolumn{4}{c|}{CHAMELEON} & \multicolumn{4}{c|}{CAMO}     & \multicolumn{4}{c|}{COD10K}   & \multicolumn{4}{c}{NC4K}      \\
                         &                       & \cellcolor{magenta!13} IoU $\uparrow$   & \cellcolor{magenta!13} Dice $\uparrow$  & \cellcolor{magenta!13}$\mathcal{F} _{m}^{w}$ $\uparrow$  & \cellcolor{magenta!13}$\mathcal{M}$ $\downarrow$   & \cellcolor{magenta!13}IoU $\uparrow$ & \cellcolor{magenta!13}Dice $\uparrow$  & \cellcolor{magenta!13}$\mathcal{F} _{m}^{w}$ $\uparrow$  & \cellcolor{magenta!13}$\mathcal{M}$ $\downarrow$   & \cellcolor{magenta!13}IoU $\uparrow$  & \cellcolor{magenta!13}Dice $\uparrow$  & \cellcolor{magenta!13}$\mathcal{F} _{m}^{w}$ $\uparrow$  & \cellcolor{magenta!13}$\mathcal{M}$ $\downarrow$   & \cellcolor{magenta!13}IoU $\uparrow$  & \cellcolor{magenta!13}Dice $\uparrow$  & \cellcolor{magenta!13}$\mathcal{F} _{m}^{w}$ $\uparrow$  & \cellcolor{magenta!13}$\mathcal{M}$ $\downarrow$   \\ \hline \hline
PFNet$_{21}$ \cite{PFNet}                   & CVPR              & 0.751  & 0.831 & 0.810 & 0.033 & 0.611 & 0.721 & 0.695 & 0.085 & 0.588 & 0.697 & 0.660 & 0.040 & 0.670 & 0.769 & 0.745 & 0.053 \\ 
JSOCOD$_{21}$  \cite{JSOCOD}                 & CVPR               & 0.776  & 0.849 & 0.833 & 0.030 & 0.649 & 0.750 & 0.728 & 0.073 & 0.612 & 0.714 & 0.684 & 0.035 & 0.698 & 0.789 & 0.771 & 0.047 \\ 
ZoomNet$_{22}$ \cite{ZoomNet}                 & CVPR               & 0.785  & 0.856 & 0.845 & 0.023 & 0.675 & 0.773 & 0.752 & 0.066 & 0.656 & 0.749 & 0.729 & 0.029 & 0.714 & 0.800 & 0.784 & 0.043 \\ 
SegMaR$_{22}$ \cite{SegMaR}                  & CVPR              & 0.804  & 0.871 & 0.860 & 0.025 & 0.675 & 0.773 & 0.753 & 0.071 & 0.656 & 0.753 & 0.724 & 0.034 & -     & -     & -     & -     \\ 
FDNet$_{22}$ \cite{FDNet}                  & CVPR              & 0.769  & 0.855 & 0.836 & 0.027 & 0.702 & 0.801 & 0.775 & 0.063 & 0.651 & 0.759 & 0.730 & 0.030 & 0.673 & 0.774 & 0.750 & 0.052 \\ 
SAM$_{23}$ \cite{SAM}                     & ICCV              & 0.560  & 0.647 & 0.639 & 0.081 & 0.522 & 0.611 & 0.606 & 0.132 & 0.616 & 0.698 & 0.701 & 0.049 & 0.615 & 0.695 & 0.696 & 0.078 \\ 
PopNet$_{23}$ \cite{PopNet}                  & ICCV             & 0.824  & 0.887 & 0.875 & 0.020 & 0.666 & 0.761 & 0.744 & 0.077 & 0.690 & 0.779 & 0.757 & 0.028 & 0.734 & 0.817 & 0.802 & 0.042 \\ 
FSPNet$_{23}$ \cite{FSPNet}                  & CVPR               & 0.786  & 0.858 & 0.851 & 0.023 & 0.721 & 0.811 & 0.799 & 0.050 & 0.651 & 0.750 & 0.735 & 0.026 & 0.742 & 0.825 & 0.816 & 0.035 \\ 
FEDER$_{23}$ \cite{FEDER}                   & CVPR              & 0.775  & 0.850 & 0.834 & 0.030 & 0.660 & 0.763 & 0.738 & 0.071 & 0.640 & 0.741 & 0.716 & 0.032 & 0.713 & 0.804 & 0.789 & 0.044 \\ 
EVP$_{23}$ \cite{EVP}                     & CVPR               & 0.707  & 0.799 & 0.777 & 0.038 & 0.674 & 0.777 & 0.762 & 0.067 & 0.641 & 0.748 & 0.726 & 0.032 & -     & -     & -     & -     \\ 
VSCode$_{24}$ \cite{VSCode}                  & CVPR              & -      & -     & -     & -     & 0.757 & 0.843 & 0.820 & 0.046 & 0.711 & 0.801 & 0.780 & 0.023 & 0.778 & 0.854 & 0.841 & 0.032 \\  
FSEL$_{24}$ \cite{FSEL}                    & ECCV               & 0.825  & \textbf{\color{myblue}0.893} & 0.877 & 0.022 & 0.792 & \textbf{\color{myblue}0.872} & 0.851 & \textbf{\color{myTeal}0.040} & 0.735 & 0.822 & 0.800 & 0.021 & 0.792 & 0.866 & 0.853 & 0.030 \\ 
CamoFormer$_{24}$ \cite{Camoformer}              & TPAMI              & 0.805  & 0.877 & 0.865 & 0.022 & 0.768 & 0.851 & 0.831 & 0.046 & 0.715 & 0.805 & 0.786 & 0.023 & 0.784 & 0.859 & 0.847 & 0.030 \\ 
ZoomXNet$_{24}$ \cite{ZoomXNet}                & TPAMI              & \textbf{\color{myblue}0.829}  & 0.891 & \textbf{\color{myblue}0.885} & \textbf{\color{myblue}0.018} & \textbf{\color{myblue}0.797} & 0.869 & \textbf{\color{myblue}0.857} & \textbf{\color{myblue}0.041} & \textbf{\color{myblue}0.758} & \textbf{\color{myblue}0.839} & \textbf{\color{myblue}0.827} & \textbf{\color{myblue}0.018} & \textbf{\color{myblue}0.799} & \textbf{\color{myblue}0.870} & \textbf{\color{myblue}0.863} & \textbf{\color{myblue}0.028} \\ \hdashline
\rowcolor{cyan!5}\textbf{Ours\_u}                  & -                     & \textbf{\color{myTeal}0.856}  & \textbf{\color{myTeal}0.913} & \textbf{\color{myTeal}0.908} & \textbf{\color{myTeal}0.016} & \textbf{\color{myTeal}0.825} & \textbf{\color{myTeal}0.893} & \textbf{\color{myTeal}0.875} & \textbf{\color{myOrange}0.035} & \textbf{\color{myTeal}0.795} & \textbf{\color{myTeal}0.866} & \textbf{\color{myTeal}0.858} & \textbf{\color{myTeal}0.015} & \textbf{\color{myTeal}0.826} & \textbf{\color{myTeal}0.887} & \textbf{\color{myTeal}0.881} & \textbf{\color{myTeal}0.024} \\ 
\rowcolor{cyan!5}\textbf{Ours\_b}                  & -                     & \textbf{\color{myOrange}0.863}  & \textbf{\color{myOrange}0.917} & \textbf{\color{myOrange}0.913} & \textbf{\color{myOrange}0.015} & \textbf{\color{myOrange}0.834} & \textbf{\color{myOrange}0.899} & \textbf{\color{myOrange}0.883} & \textbf{\color{myOrange}0.035} & \textbf{\color{myOrange}0.817} & \textbf{\color{myOrange}0.883} & \textbf{\color{myOrange}0.876} & \textbf{\color{myOrange}0.014} & \textbf{\color{myOrange}0.842} & \textbf{\color{myOrange}0.901} & \textbf{\color{myOrange}0.896} & \textbf{\color{myOrange}0.022} \\ \hline \hline
\end{tabular}}
\caption{Comparison with state-of-the-art methods on three camouflaged object detection datasets. The top three results are highlighted in \textbf{\color{myOrange} orange}, \textbf{\color{myTeal} teal}, and \textbf{\color{myblue}blue}. ``\textbf{Ours\_u}'' and ``\textbf{Ours\_b}'' denotes the fine-tuning of different frameworks, $i.e.$, UniPerceiver \cite{uniper} and BEiT \cite{Beit}.
}
\label{cod_result}
\end{table*}

\begin{table*}[]
\centering
\setlength{\tabcolsep}{5pt}
\renewcommand{\arraystretch}{0.9}
\resizebox*{\textwidth}{36mm}{
\begin{tabular}{c|c|cccc|cccc|cccc|cccc}
\hline \hline
\multirow{2}{*}{Methods} & \multirow{2}{*}{Pub.} & \multicolumn{4}{c|}{PASCAL-S} & \multicolumn{4}{c|}{ECSSD}    & \multicolumn{4}{c|}{HKU-IS}   & \multicolumn{4}{c}{DUTS-TE}   \\
                        &                              & \cellcolor{magenta!13} IoU $\uparrow$   & \cellcolor{magenta!13} Dice $\uparrow$  & \cellcolor{magenta!13}$\mathcal{F} _{m}^{w}$ $\uparrow$  & \cellcolor{magenta!13}$\mathcal{M}$ $\downarrow$   & \cellcolor{magenta!13}IoU $\uparrow$ & \cellcolor{magenta!13}Dice $\uparrow$  & \cellcolor{magenta!13}$\mathcal{F} _{m}^{w}$ $\uparrow$  & \cellcolor{magenta!13}$\mathcal{M}$ $\downarrow$   & \cellcolor{magenta!13}IoU $\uparrow$  & \cellcolor{magenta!13}Dice $\uparrow$  & \cellcolor{magenta!13}$\mathcal{F} _{m}^{w}$ $\uparrow$  & \cellcolor{magenta!13}$\mathcal{M}$ $\downarrow$   & \cellcolor{magenta!13}IoU $\uparrow$  & \cellcolor{magenta!13}Dice $\uparrow$  & \cellcolor{magenta!13}$\mathcal{F} _{m}^{w}$ $\uparrow$  & \cellcolor{magenta!13}$\mathcal{M}$ $\downarrow$   \\ \hline \hline
MENet$_{23}$ \cite{MENet}                  & CVPR                      & 0.797 & 0.865 & 0.844 & 0.054 & 0.881 & 0.924 & 0.920 & 0.031 & 0.872 & 0.922 & 0.917 & 0.023 & 0.817 & 0.880 & 0.870 & 0.028 \\ 
ICON$_{23}$ \cite{icon}                 & TPAMI                     & 0.810 & 0.877 & 0.853 & 0.051 & 0.899 & 0.939 & 0.933 & \textbf{\color{myblue}0.024} & 0.882 & 0.931 & 0.925 & 0.022 & 0.833 & 0.896 & 0.882 & \textbf{\color{myblue}0.026} \\ 
GPONet$_{24}$ \cite{GPONet}                 & PR                        & 0.797 & 0.868 & 0.845 & 0.054 & 0.894 & 0.937 & 0.932 & 0.025 & 0.869 & 0.922 & 0.918 & 0.023 & 0.817 & 0.883 & 0.872 & 0.028 \\ 
MDSAM$_{24}$ \cite{MDSAM}                  & MM                     & 0.812 & 0.876 & 0.857 & 0.051 & \textbf{\color{myblue}0.913} & \textbf{\color{myTeal}0.948} & \textbf{\color{myTeal}0.946} & \textbf{\color{myTeal}0.021} & \textbf{\color{myblue}0.893} & \textbf{\color{myblue}0.937} & \textbf{\color{myblue}0.935} & \textbf{\color{myblue}0.019} & {0.842} & {0.889} & \textbf{\color{myblue}0.893} & \textbf{\color{myTeal}0.024} \\ 
FSEL$_{24}$ \cite{FSEL}                 & ECCV                      & 0.797 & 0.866 & 0.838 & 0.057 & 0.894 & 0.936 & 0.928 & 0.026 & 0.866 & 0.919 & 0.909 & 0.027 & 0.800 & 0.866 & 0.847 & 0.037 \\ 
VSCode$_{24}$ \cite{VSCode}               & CVPR                      & \textbf{\color{myblue}0.815} & \textbf{\color{myblue}0.878} & \textbf{\color{myblue}0.859} & \textbf{\color{myblue}0.050} & 0.910 & 0.946 & 0.942 & \textbf{\color{myTeal}0.021} & 0.886 & 0.933 & 0.930 & 0.021 & \textbf{\color{myblue}0.847} & \textbf{\color{myblue}0.904} & \textbf{\color{myTeal}0.896} & \textbf{\color{myTeal}0.024} \\
VST++$_{24}$ \cite{VST++}                  & TPAMI                     & 0.801 & 0.870 & 0.846 & 0.054 & 0.890 & 0.934 & 0.926 & 0.026 & 0.867 & 0.921 & 0.914 & 0.025 & 0.810 & 0.878 & 0.866 & 0.029 \\ \hdashline
\rowcolor{cyan!5} \textbf{Ours\_u}                  & -                            & \textbf{\color{myOrange}0.842} & \textbf{\color{myOrange}0.898} & \textbf{\color{myOrange}0.882} & \textbf{\color{myOrange}0.041} & \textbf{\color{myTeal}0.917} & \textbf{\color{myblue}0.947} & \textbf{\color{myblue}0.944} & \textbf{\color{myTeal}0.021} & \textbf{\color{myOrange}0.909} & \textbf{\color{myOrange}0.946} & \textbf{\color{myOrange}0.945} & \textbf{\color{myOrange}0.016} & \textbf{\color{myOrange}0.870} & \textbf{\color{myOrange}0.916} & \textbf{\color{myOrange}0.900} & \textbf{\color{myOrange}0.021} \\ 
\rowcolor{cyan!5} \textbf{Ours\_b}                  & -                            & \textbf{\color{myTeal}0.840} & \textbf{\color{myTeal}0.897} & \textbf{\color{myTeal}0.879} & \textbf{\color{myTeal}0.043} & \textbf{\color{myOrange}0.927} & \textbf{\color{myOrange}0.955} & \textbf{\color{myOrange}0.953} & \textbf{\color{myOrange}0.017} & \textbf{\color{myTeal}0.906} & \textbf{\color{myTeal}0.945} & \textbf{\color{myTeal}0.943} & \textbf{\color{myTeal}0.017} & \textbf{\color{myTeal}0.861} & \textbf{\color{myTeal}0.908} & \textbf{\color{myOrange}0.900} & \textbf{\color{myTeal}0.024} \\ \hline \hline
\end{tabular}}
\caption{Comparison with state-of-the-art methods on four salient object detection datasets.}
\label{sod}
\end{table*}

\begin{table*}[]
\centering
\begin{minipage}[t]{\textwidth}
\centering
\setlength{\tabcolsep}{4pt}
\renewcommand{\arraystretch}{0.7}
\resizebox*{\textwidth}{31mm}{
\begin{tabular}{c|c|cccc|cccc|cccc|cccc|cccc}
\hline \hline
                                          &                        & \multicolumn{4}{c|}{CVC-300}  & \multicolumn{4}{c|}{CVC-ClinicDB} & \multicolumn{4}{c|}{Kvasir}   & \multicolumn{4}{c|}{ISIC17}   & \multicolumn{4}{c}{ISIC18}    \\
\multirow{-2}{*}{Methods}                 & \multirow{-2}{*}{Pub.} & \cellcolor{magenta!13} IoU $\uparrow$   & \cellcolor{magenta!13} Dice $\uparrow$  & \cellcolor{magenta!13}$\mathcal{F} _{m}^{w}$ $\uparrow$  & \cellcolor{magenta!13}$\mathcal{M}$ $\downarrow$   & \cellcolor{magenta!13}IoU $\uparrow$ & \cellcolor{magenta!13}Dice $\uparrow$  & \cellcolor{magenta!13}$\mathcal{F} _{m}^{w}$ $\uparrow$  & \cellcolor{magenta!13}$\mathcal{M}$ $\downarrow$   & \cellcolor{magenta!13}IoU $\uparrow$  & \cellcolor{magenta!13}Dice $\uparrow$  & \cellcolor{magenta!13}$\mathcal{F} _{m}^{w}$ $\uparrow$  & \cellcolor{magenta!13}$\mathcal{M}$ $\downarrow$   & \cellcolor{magenta!13}IoU $\uparrow$  & \cellcolor{magenta!13}Dice $\uparrow$  & \cellcolor{magenta!13}$\mathcal{F} _{m}^{w}$ $\uparrow$  & \cellcolor{magenta!13}$\mathcal{M}$ $\downarrow$  & \cellcolor{magenta!13}IoU $\uparrow$  & \cellcolor{magenta!13}Dice $\uparrow$  & \cellcolor{magenta!13}$\mathcal{F} _{m}^{w}$ $\uparrow$  & \cellcolor{magenta!13}$\mathcal{M}$ $\downarrow$   \\ \hline \hline
{PraNet$_{20}$} \cite{PraNet11}             & MICCAI                 & 0.797 & 0.871 & 0.843 & 0.010 & 0.849  & 0.899  & 0.896  & 0.009  & 0.840 & 0.898 & 0.885 & 0.030 & 0.776 & 0.853 & 0.830 & 0.050 & 0.806 & 0.881 & 0.864 & 0.057 \\ 
{DCRNet$_{22}$} \cite{DCRNet}            & ISBI                 & 0.788 & 0.856 & 0.830 & 0.010 & 0.844  & 0.896  & 0.890  & 0.010  & 0.825 & 0.886 & 0.868 & 0.035 & 0.789 & 0.866 & 0.847 & 0.046 & 0.802 & 0.876 & 0.854 & 0.059 \\ 
{CFANet$_{23}$} \cite{CFANet}             & PR                     & \textbf{\color{myblue}0.827} & \textbf{\color{myblue}0.893} & \textbf{\color{myblue}0.875} & \textbf{\color{myblue}0.008} & 0.883  & \textbf{\color{myTeal}0.932}  & 0.924  & \textbf{\color{myTeal}0.007}  & 0.861 & \textbf{\color{myTeal}0.915} & 0.903 & \textbf{\color{myblue}0.023} & 0.793 & 0.844 & 0.815 & 0.051 & 0.809 & 0.868 & 0.846 & 0.061 \\ 
{LSSNet$_{24}$} \cite{LSSNet} & MICCAI                 & 0.815 & 0.884 & 0.852 & 0.009 & 0.875  & 0.920  & 0.914  & 0.010  & \textbf{\color{myblue}0.866} & \textbf{\color{myblue}0.911} & 0.895 & 0.028 & \textbf{\color{myblue}0.813} & 0.881 & 0.867 & 0.038 & \textbf{\color{myblue}0.824} & \textbf{\color{myblue}0.886} & 0.867 & 0.054 \\ 
{LBUNet$_{24}$} \cite{LBUNet}            & MICCAI                   & 0.680 & 0.785 & 0.734 & 0.019 & 0.713  & 0.797  & 0.855  & 0.029  & 0.748 & 0.831 & 0.805 & 0.048 & 0.800 & 0.872 & 0.864 & 0.038 & 0.803 & 0.879 & 0.866 & 0.055 \\
{MEGANet$_{24}$}  \cite{MEGANet}           & WACV                   & 0.818 & 0.887 & 0.863 & 0.009 & \textbf{\color{myblue}0.885}  & \textbf{\color{myblue}0.930}  & \textbf{\color{myblue}0.931}  & \textbf{\color{myblue}0.008}  & 0.859 & \textbf{\color{myblue}0.911} & \textbf{\color{myblue}0.904} & 0.026 & 0.800 & 0.878 & 0.864 & 0.039 & 0.809 & 0.885 & 0.873 & \textbf{\color{myblue}0.052} \\ 
{FSEL$_{24}$} \cite{FSEL}            & ECCV                   & 0.814 & 0.880 & 0.856 & 0.009 & 0.867  & 0.914  & 0.910  & 0.011  & 0.852 & 0.899 & 0.894 & 0.027 & \textbf{\color{myblue}0.813} & \textbf{\color{myblue}0.885} & \textbf{\color{myblue}0.871} & \textbf{\color{myblue}0.035} & \textbf{\color{myTeal}0.821} & 0.885 & \textbf{\color{myblue}0.875} & \textbf{\color{myblue}0.052} \\ \hdashline
\rowcolor{cyan!5} \textbf{Ours\_u}                                   & -                      & \textbf{\color{myOrange}0.844} & \textbf{\color{myOrange}0.910} & \textbf{\color{myOrange}0.897} & \textbf{\color{myOrange}0.005} & \textbf{\color{myTeal}0.887}  & \textbf{\color{myblue}0.930}  & \textbf{\color{myTeal}0.932}  & \textbf{\color{myOrange}0.006}  & \textbf{\color{myTeal}0.870} & \textbf{\color{myTeal}0.915} & \textbf{\color{myTeal}0.914} & \textbf{\color{myTeal}0.021} & \textbf{\color{myOrange}0.820} & \textbf{\color{myOrange}0.890} & \textbf{\color{myOrange}0.885} & \textbf{\color{myOrange}0.032} & \textbf{\color{myOrange}0.827} & \textbf{\color{myOrange}0.897} & \textbf{\color{myOrange}0.886} & \textbf{\color{myOrange}0.045} \\
\rowcolor{cyan!5} \textbf{Ours\_b}                                   & -                      & \textbf{\color{myTeal}0.839} & \textbf{\color{myTeal}0.904} & \textbf{\color{myTeal}0.888} & \textbf{\color{myTeal}0.006} & \textbf{\color{myOrange}0.896}  & \textbf{\color{myOrange}0.935}  & \textbf{\color{myOrange}0.939}  & \textbf{\color{myOrange}0.006}  & \textbf{\color{myOrange}0.885} & \textbf{\color{myOrange}0.930} & \textbf{\color{myOrange}0.928} & \textbf{\color{myOrange}0.017} & \textbf{\color{myTeal}0.814} & \textbf{\color{myTeal}0.887} & \textbf{\color{myTeal}0.878} & \textbf{\color{myTeal}0.034} & 0.817 & \textbf{\color{myTeal}0.889} & \textbf{\color{myTeal}0.878} & \textbf{\color{myTeal}0.049} \\  \hline \hline
\end{tabular}}
\caption{Comparison with state-of-the-art methods on three polyp segmentation and two skin lesion segmentation datasets.}
\label{MIS_results}
\end{minipage}
\hfill

\begin{minipage}[t]{0.50\textwidth}
\centering
\setlength{\tabcolsep}{2pt}
\renewcommand{\arraystretch}{0.7}
\resizebox*{\textwidth}{22mm}{
\begin{tabular}{c|c|cccc|cccc|cccc}
\hline \hline
\multirow{2}{*}{Methods} & \multirow{2}{*}{Pub.} & \multicolumn{4}{c|}{SBU}      & \multicolumn{4}{c|}{UCF}      & \multicolumn{4}{c}{ISTD}      \\
                         &                       & \cellcolor{magenta!13}IoU $\uparrow$  & \cellcolor{magenta!13}Dice $\uparrow$  & \cellcolor{magenta!13}$\mathcal{F} _{m}^{w}$ $\uparrow$  & \cellcolor{magenta!13}$\mathcal{M}$ $\downarrow$   & \cellcolor{magenta!13}IoU $\uparrow$  & \cellcolor{magenta!13}Dice $\uparrow$  & \cellcolor{magenta!13}$\mathcal{F} _{m}^{w}$ $\uparrow$  & \cellcolor{magenta!13}$\mathcal{M}$ $\downarrow$   & \cellcolor{magenta!13}IoU $\uparrow$  & \cellcolor{magenta!13}Dice $\uparrow$  & \cellcolor{magenta!13}$\mathcal{F} _{m}^{w}$ $\uparrow$  & \cellcolor{magenta!13}$\mathcal{M}$ $\downarrow$  \\ \hline \hline
RMLA$_{23}$ \cite{rmlanet}                 & TCSVT             & 0.798 & 0.878 & 0.839 & 0.032 & 0.669 & 0.780 & 0.713 & 0.064 & \textbf{\color{myblue}0.910} & \textbf{\color{myTeal}0.947} & \textbf{\color{myblue}0.931} & \textbf{\color{myTeal}0.012} \\ 
SDSAM$_{23}$ \cite{sdsam}               & TGRS               & 0.787 & 0.841 & 0.816 & 0.042 & \textbf{\color{myblue}0.678} & 0.746 & 0.670 & 0.071 & 0.891 & 0.921 & 0.902 & 0.023 \\ 
EVP$_{23}$ \cite{EVP}                     & CVPR               & 0.815 & 0.853 & 0.809 & 0.038 & 0.667 & 0.745 & 0.672 & 0.071 & 0.861 & 0.886 & 0.855 & 0.031 \\ 
FSEL$_{24}$ \cite{FSEL}                    & ECCV               & \textbf{\color{myblue}0.835} & \textbf{\color{myblue}0.893} & \textbf{\color{myblue}0.875} & \textbf{\color{myblue}0.028} & \textbf{\color{myTeal}0.703} & \textbf{\color{myblue}0.793} & \textbf{\color{myblue}0.745} & \textbf{\color{myblue}0.057} & 0.908 & \textbf{\color{myblue}0.945} & 0.930 & \textbf{\color{myblue}0.013} \\ 
Spider$_{24}$ \cite{Spider}                  & ICML               & 0.823 & \textbf{\color{myblue}0.893} & 0.868 & \textbf{\color{myTeal}0.027} & -     & -     & -     & -     & -     & -     & -     & -     \\ \hdashline
\rowcolor{cyan!5}\textbf{Ours\_u}                  & -                     & \textbf{\color{myTeal}0.857} & \textbf{\color{myTeal}0.914} & \textbf{\color{myTeal}0.909} & \textbf{\color{myOrange}0.022} & \textbf{\color{myOrange}0.736} & \textbf{\color{myOrange}0.831} & \textbf{\color{myOrange}0.797} & \textbf{\color{myOrange}0.042} & \textbf{\color{myOrange}0.941} & \textbf{\color{myOrange}0.965} & \textbf{\color{myOrange}0.959} & \textbf{\color{myOrange}0.008} \\ 
\rowcolor{cyan!5}\textbf{Ours\_b}                  & -                     & \textbf{\color{myOrange}0.861} & \textbf{\color{myOrange}0.917} & \textbf{\color{myOrange}0.912} & \textbf{\color{myOrange}0.022} & \textbf{\color{myOrange}0.736} & \textbf{\color{myTeal}0.828} & \textbf{\color{myTeal}0.795} & \textbf{\color{myTeal}0.043} & \textbf{\color{myTeal}0.916} & \textbf{\color{myTeal}0.947} & \textbf{\color{myTeal}0.935} & 0.015 \\ \hline \hline
\end{tabular}}
\caption{Comparison with recent state-of-the-art methods on three shadow detection datasets.}
\label{SD_results}
\end{minipage}
\hfill
\begin{minipage}[t]{0.48\textwidth}
\centering
\setlength{\tabcolsep}{4.5pt}
\renewcommand{\arraystretch}{0.5}
\resizebox*{\textwidth}{22mm}{
\begin{tabular}{c|c|cccc|cccc}
\hline \hline
\multirow{2}{*}{Methods} & \multirow{2}{*}{Pub.} & \multicolumn{4}{c|}{Trans10k} & \multicolumn{4}{c}{GDD}       \\
                         &                       & \cellcolor{magenta!13}IoU $\uparrow$  & \cellcolor{magenta!13}Dice $\uparrow$  & \cellcolor{magenta!13}$\mathcal{F} _{m}^{w}$ $\uparrow$  & \cellcolor{magenta!13}$\mathcal{M}$ $\downarrow$   & \cellcolor{magenta!13}IoU $\uparrow$  & \cellcolor{magenta!13}Dice $\uparrow$  & \cellcolor{magenta!13}$\mathcal{F} _{m}^{w}$ $\uparrow$  & \cellcolor{magenta!13}$\mathcal{M}$ $\downarrow$   \\ \hline \hline
EBLNet$_{21}$ \cite{EBLNet}                  & ICCV               & 0.888 & 0.934 & 0.911 & \textbf{\color{myblue}0.044} & 0.884 & 0.929 & 0.910 & 0.055 \\ 
GlassNet$_{22}$ \cite{GlassNet}                & NeurIPS            & 0.838 & 0.903 & 0.860 & 0.067 & -     & -     & -     & -     \\ 
ICON$_{23}$ \cite{icon}                    & TPAMI              & 0.889 & 0.930 & 0.906 & 0.046 & 0.900 & \textbf{\color{myblue}0.937} & 0.917 & 0.051 \\ 
FSPNet$_{23}$ \cite{FSPNet}                  & CVPR               & \textbf{\color{myblue}0.896} & 0.934 & 0.914 & \textbf{\color{myTeal}0.043} & 0.903 & \textbf{\color{myblue}0.937} & 0.921 & 0.049 \\ 
RFENet$_{23}$ \cite{RFENet}                  & IJCAI              & 0.892 & \textbf{\color{myblue}0.937} & \textbf{\color{myblue}0.915} & \textbf{\color{myTeal}0.043} & 0.871 & 0.919 & 0.897 & 0.061 \\ 
FSEL$_{24}$ \cite{FSEL}                    & ECCV               & 0.892 & 0.934 & 0.913 & \textbf{\color{myTeal}0.043} & \textbf{\color{myblue}0.906} & \textbf{\color{myTeal}0.942} & \textbf{\color{myblue}0.924} & \textbf{\color{myblue}0.047} \\ \hdashline
\rowcolor{cyan!5}\textbf{Ours-u}                   & -                     & \textbf{\color{myTeal}0.930} & \textbf{\color{myTeal}0.960} & \textbf{\color{myTeal}0.947} & \textbf{\color{myOrange}0.027} & \textbf{\color{myOrange}0.923} & \textbf{\color{myOrange}0.952} & \textbf{\color{myOrange}0.941} & \textbf{\color{myTeal}0.039} \\ 
\rowcolor{cyan!5}\textbf{Ours-b}                   &   -                    & \textbf{\color{myOrange}0.931} & \textbf{\color{myOrange}0.961} & \textbf{\color{myOrange}0.948} & \textbf{\color{myOrange}0.027} & \textbf{\color{myTeal}0.922} & \textbf{\color{myOrange}0.952} & \textbf{\color{myTeal}0.940} & \textbf{\color{myOrange}0.037} \\ \hline \hline
\end{tabular}}
\caption{Comparison with recent state-of-the-art methods on two glass detection datasets.}
\label{GD_results}
\end{minipage}
\end{table*}

\textbf{Channel-oriented adaptive scale enhancement.} The optimized feature $\hat{f}_u^{1}$ is input into the $1$-th encoding block to obtain the output feature ${f}_u^{2}$. Then, we conduct the second knowledge exchange to enhance the expressive ability of the feature $f_{ts}^1$. In detail, as opposed to the first exchange, we take the specific feature $f_{ts}^1$ as the primary $query$, and the output feature $f_{u}^2$ as an auxiliary $value$ into our CDA component for interactive fusion, that is,
\begin{equation}
\begin{split}
&\hat{f}_{ts}^{1}=f_{ts}^{1}+\mathsf{CDA}(\mathsf{Norm}(f_{ts}^{1}),\mathsf{Norm}(f_{u}^{2})), \\
\end{split}
\end{equation}
where $\mathsf{CDA}(\cdot)$ denotes the proposed CDA mechanism. Furthermore, we construct the CASE to strengthen multi-scale information with the channels to generate the task-specific feature $f_{ts}^2$ for the next stage of interaction. Technically, we first reinterpret the input feature $\hat{f}_{ts}^{1}$ by decomposing it into three features and enhancing its linear expression through the depthwise separable convolution \cite{DSC} with the $3 \times 3$ kernal ($\mathcal{DC}_3(\cdot)$), $i.e.$, $(\check{f}_{ts}^1)_1,(\check{f}_{ts}^1)_2,(\check{f}_{ts}^1)_3 =\mathcal{DC}_3(\mathsf{Split}(\mathsf{Norm}(\hat{f}_{ts}^1)))$. Then, we dynamically regulate significant clues within the channel from two perspectives through the channel and reverse attentions \cite{RA,CA}. Similarly to our DMLP extractor, we regard the outputs from two perspectives as two experts and adaptively fuse them using dynamic weights $w_t^x$ ($w_t^x= \mathsf{Softmax}(\mathsf{W}_g \hat{f}_{ts}^{1}+\mathsf{b}_g)$) generated by a gating network. The process is as follows: 
\begin{equation}
\begin{split}
&f_{ts}^2=\hat{f}_{ts}^{1}+\mathsf{flat}([\hat{f}_s^2,\hat{f}_s^3,\hat{f}_s^{4}]), \hat{f}_s^k=\sum_{x=1}^{X}w_{t}^{x}\otimes(\mathrm{E}_c^x)_k, \\
&(\mathrm{E}_c^1)_k,(\mathrm{E}_c^2)_k= \mathsf{CA}((\check{f}_{ts}^1)_k), \mathsf{RA}((\check{f}_{ts}^1)_k), k=2,3,4, \\
\end{split}
\end{equation}
where $\mathsf{flat}(\cdot)$ is a flattening operation, $\mathsf{CA}(\cdot)$ and $\mathsf{RA}(\cdot)$ represent the channel \cite{CA} and reverse \cite{RA} attentions, respectively. Similarly, the obtained features $f_{ts}^2$ and $f_{u}^2$ interactively fuse in the $2$-th block to generate the features $f_{ts}^3$ and $f_{u}^3$. The entire fine-tuning continues until the 4-th block.

\subsection{Loss functions}
After the interaction is completed, the optimized features of each block are input into a lightweight Transformer-based mask decoder \cite{mask2}, which contains 1.08M parameters, for decoding and output. During the fine-tuning process, we use the binary cross-entropy loss and the Dice coefficient loss to supervise the training of our model, as follows: 
\begin{equation}
\begin{split}
&\mathcal{L}_{all}=\alpha\mathcal{L}_{bce}+\beta\mathcal{L}_{dice}, \\
\end{split}
\end{equation}
where $\alpha$ and $\beta$ represent the hyperparameters set to 5 and 2.

\begin{figure*}[]
	\centering\includegraphics[width=0.97\textwidth,height=7.6cm]{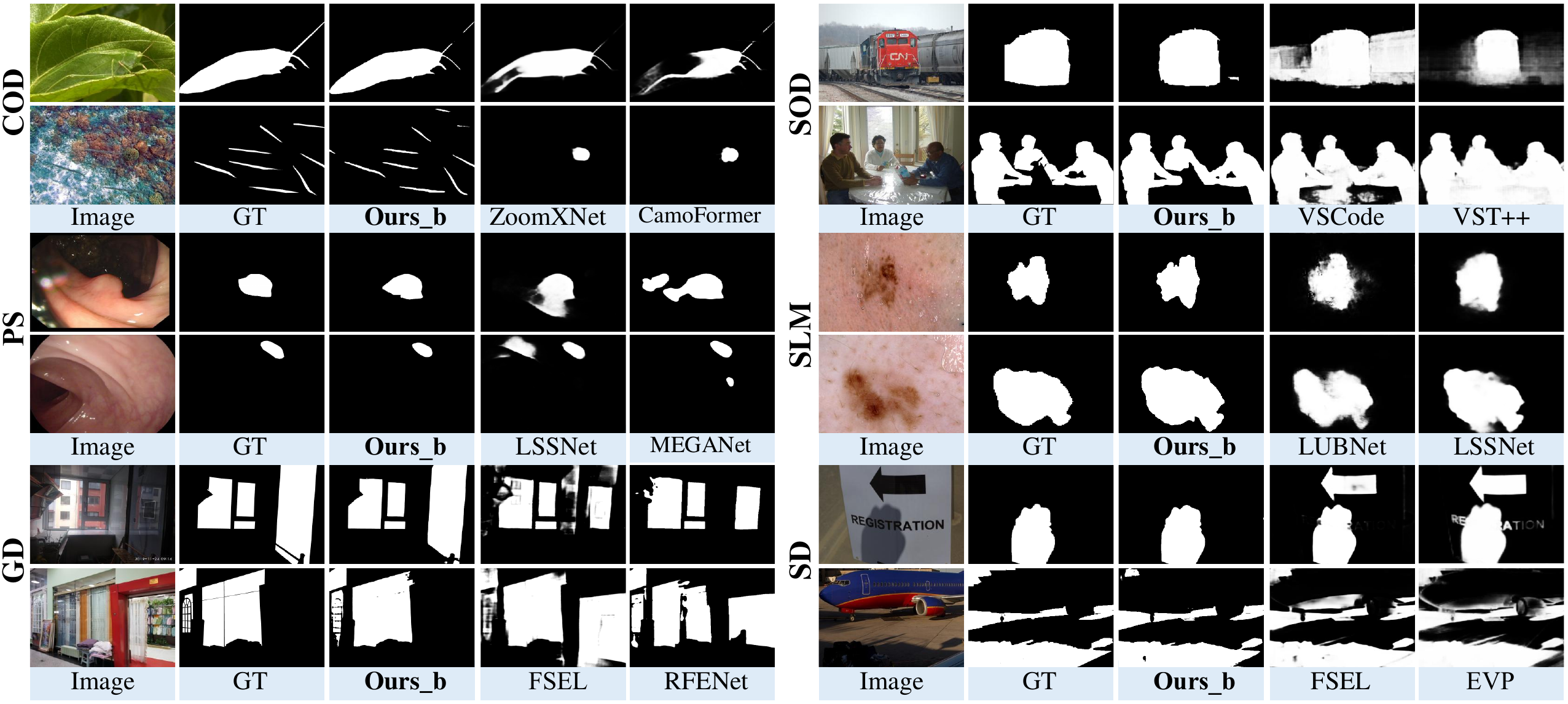}
	\caption{Visual comparison results of our Controllable-LPMoE with multiple state-of-the-art methods on six binary segmentation tasks.}
	\label{visual_results}
\end{figure*}
\section{Experiment}
\subsection{Experimental Settings}
\textbf{Datasets.} We evaluate our Controllable-LPMoE method on multiple binary object segmentation tasks, including camouflaged object detection (COD), salient object detection (SOD), polyp segmentation (PS), skin lesion segmentation (SLS), shadow detection (SD), and glass detection (GD). For COD, we utilize CAMO-TR\cite{CAMO} and COD10K-TR \cite{COD10K} as joint training datasets and evaluate the accuracy in CHAMELEON \cite{CHAMELEON}, CAMO-TE \cite{CAMO}, COD10K-TE \cite{COD10K}, and NC4K \cite{NC4K} datasets. In SOD, DUTS-TR \cite{DUTS} is employed as the training dataset, while performance is assessed on PASCAL-S \cite{PASCAL-S}, ECSSD \cite{ECSSD}, HKU-IS \cite{HKU}, and DUTS-TE \cite{DUTS}. Regarding PS, we use training images from CVC-ClinicDB \cite{cvc} and Kvasir \cite{kvasir} to train the model and validate its performance on test images from CVC-300 \cite{CVC300}, CVC-ClinicDB \cite{cvc}, and Kvasir \cite{kvasir}. For SLS, we train/test performance in the ISIC17 \cite{ISIC17} and ISIC18 \cite{ISIC18} datasets, respectively. In SD, we train the model on the training images of SBU \cite{SBU} and ISTD \cite{ISTD} and evaluate its performance of UCF \cite{UCF}, SBU \cite{SBU}, and ISTD \cite{ISTD}. In addition, we use Trans10k \cite{Trans10k} and GDD \cite{GDD} datasets for both training and testing in the GD task. More details of the datasets are presented in the \textbf{supplementary materials}.

\textbf{Implementation details.} All experiments are conducted on four NVIDIA GTX 4090 GPUs, each equipped with 24GB of memory. We utilize the frozen BEiT-L \cite{Beit} and UniPerceiver-L \cite{uniper} frameworks, adapting them for binary segmentation tasks through efficient fine-tuning. During fine-tuning, input images are resized to 512 × 512, the batch size is set to 4, and the initial learning rate is 5e-5. The entire training process runs for 80K iterations, with the proposed model optimized using the AdamW optimizer.

\textbf{Evaluation metrics.} We use four evaluation metrics to verify the superiority of our model, including mean Intersection over Union (IoU), mean Dice Coefficient (Dice), weighted F-measure ($\mathcal{F} _{m}^{w}$), and mean absolute error ($\mathcal{M}$). Better segmentation results are indicated by larger scores for IoU, Dice, and $\mathcal{F} _{m}^{w}$, along with a smaller $\mathcal{M}$ value.

\subsection{Comparison with the State-of-the-Art}
We compare the performance of the Controllable-LPMoE model against 31 state-of-the-art methods from six different binary object segmentation tasks. In particular, the prediction maps for all competing methods are either provided directly by their respective authors or obtained by training their publicly available open-source code.

\textbf{Quantitative evaluation.} Tables \ref{cod_result}-\ref{GD_results} present the quantitative results of our model and 31 existing segmentation approaches. From Table \ref{cod_result}, the ``IoU'' metric has improved in four widely utilized COD datasets, increasing by 4.10\%, 4.64\%, 7.78\%, and 5.38\% over the recent ZoomXNet \cite{ZoomXNet} method, and by 4.61\%, 5.30\%, 11.16\%, and 6.31\% over the recent FSEL \cite{FSEL} method in the highly challenging COD task.
For the SOD task in Table \ref{sod}, compared to the recently proposed VSCode \cite{VSCode} model, our Controllable-LPMoE model achieves overall improvements of 21.95\%, 23.53\%, 31.25\%, and 14.29\% on four public datasets in terms of the ``$\mathcal{M}$'' metric. Similarly, the proposed Controllable-LPMoE method demonstrates significant superiority across various metrics in other segmentation tasks, as detailed in Tables \ref{MIS_results}, \ref{SD_results}, and \ref{GD_results}. This performance advantage stems from the joint fine-tuning of our DMLP extractor and BDI adapter, which enables the internal features within large-scale models to be efficiently adapted for binary segmentation tasks.

\textbf{Qualitative evaluation.} Fig. \ref{visual_results} illustrates visual comparison results in various scenarios. As depicted in Figure \ref{visual_results}, the proposed Controllable-LPMoE method demonstrates superior segmentation accuracy across different objects, generating predicted maps that not only retain complete object structures, but also exhibit sharp and well-defined edge details. In contrast, some existing methods \cite{ZoomXNet,FSEL,LSSNet,Camoformer} struggle to achieve this level of precision.

\subsection{Ablation Study}
To verify the contribution of each key design and the rationale behind its internal structures, we conduct extensive ablation studies based on the UniPerceiver \cite{uniper} framework.

\begin{figure}[]
	\centering\includegraphics[width=0.48\textwidth,height=3.6cm]{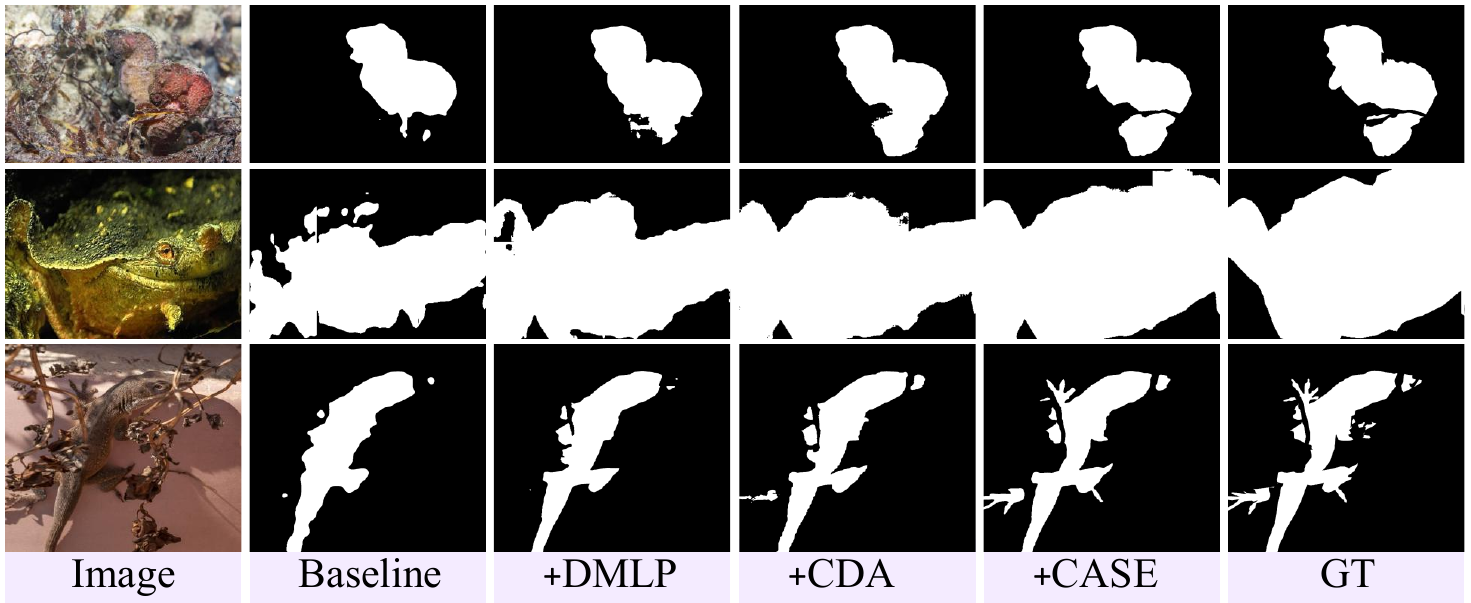}
	\caption{Visual results of the effectiveness of each component.}
	\label{ablation_results}
\end{figure}

\textbf{Effect of each component.} In Table \ref{each component}, we give the quantitative results of each component in the proposed Controllable-LPMoE method. Specifically, the ``baseline'' (Table \ref{each component}{\color{red}(a)}) includes a UniPerceiver \cite{uniper} network with frozen parameters and a mask decoder \cite{mask2}. Table \ref{each component}{\color{red}(b)}) validates the effectiveness of our ``DMLP'' extractor, demonstrating that embedding dynamic local priors with semantic aids in adapting frozen frameworks to binary object segmentation tasks. Furthermore, as shown in Tables \ref{each component} {\color{red}(c)} and {\color{red}(e)}, the proposed ``CDA'' component significantly improves segmentation accuracy through the effective interaction between frozen and trainable features. Subsequently, we incorporate the designed ``CASE'' into the BDI adapter (as shown in Table \ref{each component} {\color{red}(d)} and {\color{red}(f)}), further improving the performance by optimizing the scale information of the task-specific features and adaptively adjusting the significant clues within the channels. Furthermore, in Fig. \ref{ablation_results}, we present the visual results obtained by gradually adding each component ($i.e.$, DMLP, CDA, and CASE), demonstrating that the predicted map gradually approaches the ground truth (GT). In short, each component is necessary and collectively improves the ``baseline'' by 27.38\%, 17.02\%, 21.93\%, and 11.32\% under the ``IoU'' metric.
\begin{table*}[]
\centering
\begin{minipage}{0.48\textwidth}
\setlength{\tabcolsep}{2pt}
\renewcommand{\arraystretch}{0.7}
\resizebox*{\textwidth}{23mm}{
\begin{tabular}{c|cccc|cc|cc|cc|cc}
\toprule[1.5pt]
\multirow{2}{*}{Num.} & \multicolumn{4}{c|}{Structure Settings} & \multicolumn{2}{c|}{CHAMELEON} & \multicolumn{2}{c|}{CAMO} & \multicolumn{2}{c|}{COD10K} & \multicolumn{2}{c}{NC4K} \\
                      & Base.  & DMLP & CDA & CASE  & \cellcolor{magenta!13} IoU $\uparrow$           & \cellcolor{magenta!13}$\mathcal{F} _{m}^{w}$ $\uparrow$           & \cellcolor{magenta!13}IoU $\uparrow$           & \cellcolor{magenta!13}$\mathcal{F} _{m}^{w}$ $\uparrow$         & \cellcolor{magenta!13}IoU $\uparrow$           & \cellcolor{magenta!13}$\mathcal{F} _{m}^{w}$ $\uparrow$         & \cellcolor{magenta!13}IoU $\uparrow$           & \cellcolor{magenta!13}$\mathcal{F} _{m}^{w}$ $\uparrow$        \\ \midrule[1.5pt]
(a)                   & $\checkmark$        &         &         &         & 0.672          & 0.741         & 0.705       & 0.774       & 0.652        & 0.729        & 0.742       & 0.809      \\ 
(b)                   & $\checkmark$         & $\checkmark$       &         &         & 0.787          & 0.848         & 0.781       & 0.841       & 0.752        & 0.820        & 0.803       & 0.861      \\ 
(c)                   & $\checkmark$         &         & $\checkmark$       &         & 0.799          & 0.864         & 0.803       & 0.862       & 0.754        & 0.824        & 0.812       & 0.872      \\ 
(d)                   & $\checkmark$         &        & $\checkmark$       & $\checkmark$         & 0.816          & 0.877         & 0.804       & 0.859       & 0.763        & 0.832        & 0.817       & 0.874 
\\ 
(e)                   & $\checkmark$         & $\checkmark$       & $\checkmark$       &         & 0.839          & 0.897         & 0.821       & \textbf{0.876}       & 0.774        & 0.841        & 0.819       & 0.876 
\\ 
\rowcolor{blue!6} (f)                   & $\checkmark$         & $\checkmark$       & $\checkmark$       & $\checkmark$       & \textbf{0.856}          & \textbf{0.908}         & \textbf{0.825}       & 0.875       & \textbf{0.795}        & \textbf{0.858}        & \textbf{0.826}       & \textbf{0.881}      \\ \bottomrule[1.5pt]  
\end{tabular}}
\caption{Ablation study of individual components in the proposed Controllable-LPMoE framework on challenging COD tasks.}
\label{each component}
\end{minipage}%
\hfill
\begin{minipage}{0.48\textwidth}
\setlength{\tabcolsep}{2pt}
\renewcommand{\arraystretch}{0.7}
\resizebox*{\textwidth}{25mm}{
\begin{tabular}{c|ccccc|cc|cc|cc|cc}
\toprule[1.5pt]
\multirow{2}{*}{Num.} & \multicolumn{5}{c|}{DMLP Extractor Settings} & \multicolumn{2}{c|}{CHAMELEON} & \multicolumn{2}{c|}{CAMO} & \multicolumn{2}{c|}{COD10K} & \multicolumn{2}{c}{NC4K} \\
                      & $E1$  & $E2$  & $E3$  & $E4$  & DCS  & \cellcolor{magenta!13} IoU $\uparrow$           & \cellcolor{magenta!13}$\mathcal{F} _{m}^{w}$ $\uparrow$           & \cellcolor{magenta!13}IoU $\uparrow$           & \cellcolor{magenta!13}$\mathcal{F} _{m}^{w}$ $\uparrow$         & \cellcolor{magenta!13}IoU $\uparrow$           & \cellcolor{magenta!13}$\mathcal{F} _{m}^{w}$ $\uparrow$         & \cellcolor{magenta!13}IoU $\uparrow$           &\cellcolor{magenta!13} $\mathcal{F} _{m}^{w}$ $\uparrow$        \\ \midrule[1pt]
(a)                   & $\checkmark$   &     &     &     &      & 0.741          & 0.808         & 0.757       & 0.821       & 0.712        & 0.786        & 0.786       & 0.847      \\ 
(b)                   &     & $\checkmark$   &     &     &      & 0.741          & 0.806         & 0.760       & 0.820       & 0.711        & 0.783        & 0.784       & 0.844      \\ 
(c)                   &     &     & $\checkmark$   &     &      & 0.730          & 0.800         & 0.750       & 0.812       & 0.707        & 0.780        & 0.781       & 0.842      \\ 
(d)                   &     &     &     & $\checkmark$   &      & 0.734          & 0.802         & 0.754       & 0.816       & 0.713        & 0.786        & 0.786       & 0.846      \\ 
(e)                   & $\checkmark$    &  $\checkmark$   &     &    &  $\checkmark$    & 0.755          & 0.821         & 0.765       & 0.832       & 0.714        & 0.791        & 0.790       & 0.852      \\ 
(f)                   & $\checkmark$     & $\checkmark$     &  $\checkmark$   &    &  $\checkmark$    & 0.767          & 0.828         & 0.765       & 0.830       & 0.730        & 0.800        & 0.794       & 0.852      \\ 
(g)                   & $\checkmark$   & $\checkmark$   & $\checkmark$   & $\checkmark$   &      & 0.757          & 0.824         & 0.764       & 0.825       & 0.720        & 0.793        & 0.790       & 0.850      \\ 
\rowcolor{blue!6}(h)                   & $\checkmark$   & $\checkmark$   & $\checkmark$   & $\checkmark$   & $\checkmark$    &\textbf{0.787}          & \textbf{0.848}         & \textbf{0.781}       & \textbf{0.841}       & \textbf{0.752}        & \textbf{0.820}        & \textbf{0.803}       & \textbf{0.861}      \\ \bottomrule[1.5pt]
\end{tabular}}
\caption{Ablation study on the internal structure of our DMLP Extractor. ``$E_1$''-``$E_4$'' represent different local prior knowledge. }
\label{DMLP_e1_e4}
\end{minipage}
\end{table*}

\textbf{Effect of local priors within the DMLP extractor.} Do we really need various local priors? To answer this question, we assess the impact of each local prior in the proposed DMLP extractor (as depicted in Table \ref{DMLP_e1_e4}{\color{red}(a)}-{\color{red}(d)}). These results show that incorporating local priors enhances model performance, benefiting both from the semantic prompts they carry and the rich spatial details they provide. 
Moreover, we conduct an experimental analysis of the number of experts in Table \ref{DMLP_e1_e4} {\color{red}(e)} and {\color{red}(f)}. Furthermore, we analyze the dynamic control strategy (DCS). Table \ref{DMLP_e1_e4}{\color{red}(g)} represents the fusion of all local priors using ``element-wise addition'', while Table \ref{DMLP_e1_e4}{\color{red}(h)} denotes the aggregation through a gating network with dynamic weights, with the latter performing better. In conclusion, the design of the proposed DMLP extractor is both well-reasoned and effective.

\begin{table}[]
\centering
\setlength{\tabcolsep}{3pt}
\renewcommand{\arraystretch}{0.6}
\resizebox*{0.48\textwidth}{15mm}{
\begin{tabular}{c|cc|cc|cc|cc|cc}
\toprule[1.5pt]
\multirow{2}{*}{Num.} & \multicolumn{2}{c|}{BDI Adapter Settings} & \multicolumn{2}{c|}{CHAMELEON} & \multicolumn{2}{c|}{CAMO} & \multicolumn{2}{c|}{COD10K} & \multicolumn{2}{c}{NC4K} \\
                      & $t$$\longrightarrow $ $f$          & $f$$\longrightarrow $ $t$          & \cellcolor{magenta!13} IoU $\uparrow$           & \cellcolor{magenta!13}$\mathcal{F} _{m}^{w}$ $\uparrow$           & \cellcolor{magenta!13}IoU $\uparrow$           & \cellcolor{magenta!13}$\mathcal{F} _{m}^{w}$ $\uparrow$         & \cellcolor{magenta!13}IoU $\uparrow$           & \cellcolor{magenta!13}$\mathcal{F} _{m}^{w}$ $\uparrow$         & \cellcolor{magenta!13}IoU $\uparrow$           & \cellcolor{magenta!13}$\mathcal{F} _{m}^{w}$ $\uparrow$        \\ \midrule[1pt]
(a)                   & $\checkmark$            &              & 0.796          & 0.858         & 0.779       & 0.840       & 0.740        & 0.811        & 0.796       & 0.856      \\ 
(b)                   &              & $\checkmark$            & 0.787          & 0.848         & 0.781       & 0.841       & 0.752        & 0.820        & 0.803       & 0.861      \\ 
\rowcolor{blue!6}(c)                   & $\checkmark$            & $\checkmark$            & \textbf{0.799}          & \textbf{0.864}         & \textbf{0.803}       & \textbf{0.862}       & \textbf{0.754}        & \textbf{0.824}        & \textbf{0.812}       & \textbf{0.872}      \\ \bottomrule[1.5pt]
\end{tabular}}
\caption{Ablation study on the bi-directional interaction architecture of the proposed BDI Adapter. ``$f$'' and  ``$t$'' denote frozen features and trainable features, respectively.}
\label{bi-directional}
\end{table}

\begin{figure}[]
	\centering\includegraphics[width=0.49\textwidth,height=3.5cm]{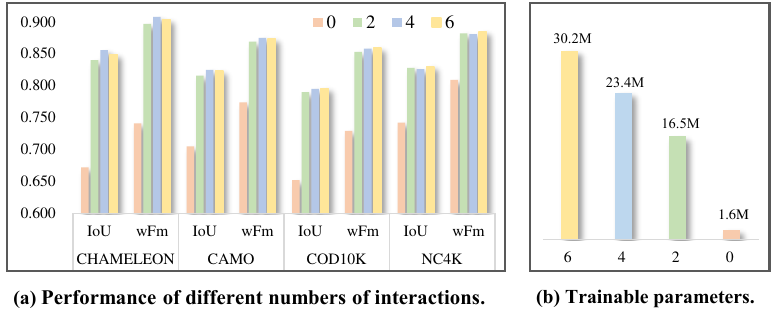}
	\caption{Ablation analysis of different interaction numbers. Here, we set the number of interactions to 0, 2, 4, and 6, respectively}
	\label{interaction numabers}
\end{figure}

\textbf{Effect of bi-directional interaction within the BDI adapter.} Bi-directional interaction aims to achieve efficient fine-tuning by leveraging frozen features to enhance the generality of trainable features. Conversely, when the focus shifts to frozen features, frozen features are imbued with task-specific attributes. Table \ref{bi-directional} {\color{red}(a)} and {\color{red}(b)} present the quantitative results for different features as subjects of interaction. Compared to the bi-directional strategy (Table \ref{bi-directional} {\color{red}(c)}), it is evident that a single interaction performs significantly worse. Furthermore, we conduct an experimental analysis on the impact of the number of interactions. As illustrated in Fig. \ref{interaction numabers}, an increase in the interaction numbers leads to a corresponding increase in the trainable parameters, which in turn enhances performance to a certain extent. To strike a balance between efficiency and performance, we set the number of interactions to 4. These results highlight the effectiveness of the proposed BDI adapter. 

\textbf{Efficiency analysis.} In Table \ref{efficient_analysis}, we present key metrics of our method under different training paradigms ($i.e.$, full-parameter fine-tuning and dynamic priors-based fine-tuning), including the trainable parameters, the memory required for training, the time consumed per 50 iterations, and the corresponding performance. From Table \ref{efficient_analysis}, compared to the full-parameter fine-tuning, the number of parameters required for training is only about 1/14, significantly reducing computational costs. Meanwhile, memory consumption during training decreased by 35.49\% and 55.79\%, while training speed improved by 10.74\% and 13.66\%. Although its performance is slightly lower than the full-parameter fine-tuning, the overall performance remains excellent. These results further demonstrate the efficiency of our dynamic priors-based fine-tuning paradigm.

\begin{table}[]
\centering
\setlength{\tabcolsep}{2pt}
\renewcommand{\arraystretch}{0.7}
\resizebox*{0.48\textwidth}{19mm}{
\begin{tabular}{c|c|c|c|cc|cc|cc|cc}
\toprule[1.5pt]
\multirow{2}{*}{Methods} & \multirow{2}{*}{\begin{tabular}[c]{@{}c@{}}Train.\\ parameters\end{tabular}} & \multirow{2}{*}{\begin{tabular}[c]{@{}c@{}}Train.\\ memory\end{tabular}} & \multirow{2}{*}{\begin{tabular}[c]{@{}c@{}}Train.\\ times\end{tabular}} & \multicolumn{2}{c|}{CHAMELEON} & \multicolumn{2}{c|}{CAMO} & \multicolumn{2}{c|}{COD10K} & \multicolumn{2}{c}{NC4K} \\
                         &                                                                                 &                                                                            &                                                                           & \cellcolor{magenta!13} IoU $\uparrow$           & \cellcolor{magenta!13}$\mathcal{F} _{m}^{w}$ $\uparrow$           & \cellcolor{magenta!13}IoU $\uparrow$           & \cellcolor{magenta!13}$\mathcal{F} _{m}^{w}$ $\uparrow$         & \cellcolor{magenta!13}IoU $\uparrow$           & \cellcolor{magenta!13}$\mathcal{F} _{m}^{w}$ $\uparrow$         & \cellcolor{magenta!13}IoU $\uparrow$           & \cellcolor{magenta!13}$\mathcal{F} _{m}^{w}$ $\uparrow$        \\ \midrule[1pt]
Ours\_u$^{\dag}$              & 326.7M                                                                          & 18.63G                                                                     & 0.907                                                                & 0.861          & 0.911         & 0.840       & 0.888       & 0.815        & 0.875        & 0.838       & 0.890      \\                          
\rowcolor{blue!6}Ours\_u               & \textbf{23.4M}                                                                           & 13.75G                                                                     & \textbf{0.819}                                                                & 0.856          & 0.908         & 0.825       & 0.875       & 0.795        & 0.858        & 0.826       & 0.881      \\ 
Ours\_b$^{\dag}$              & 328M                                                                            & 13.32G                                                                     & 1.032                                                                & \textbf{0.866}          & \textbf{0.914}         & \textbf{0.851}       & \textbf{0.898}       & \textbf{0.825}        & \textbf{0.882}        & \textbf{0.842}       & 0.895      \\
\rowcolor{blue!6}Ours\_b               & \textbf{23.4M}                                                                           & \textbf{8.55G}                                                                      & 0.908                                                                & 0.863          & 0.913         & 0.834       & 0.884       & 0.817        & 0.876        & \textbf{0.842}       & \textbf{0.896}      \\ 
\bottomrule[1.5pt]
\end{tabular}}
\caption{Efficiency analysis for our proposed method, where ``$\dag$'' indicates training with the full-parameter fine-tuning strategy.}
\label{efficient_analysis}
\end{table}

\section{Conclusion}
In this paper, we propose a novel Controllable-LPMoE method, specifically designed for fine-tuning large-scale models to adapt to binary object segmentation tasks. First, we develop a lightweight DMLP extractor, which generates task-specific features enriched with dynamic local priors, thereby providing more effective support for fine-tuning. Second, we design the BDI adapter, which facilitates efficient interaction between frozen and trainable features to update both types of information. Extensive experiments demonstrate that our method obviously surpasses 31 SOTA models in 18 binary object segmentation datasets.

\section*{Acknowledgments} This work was supported in part by the National Science Fund of China (No. 62276135, U24A20330 and 62361166670).
{
    \small
    \bibliographystyle{ieeenat_fullname}
    \bibliography{main}
}

\end{document}